\documentclass[12pt]{article}
\usepackage[a4paper,margin=1in]{geometry}
\usepackage{lineno}
\usepackage{booktabs}
\usepackage{multirow}
\usepackage{float}
\usepackage{authblk}
\usepackage{amsmath}
\usepackage[pdftex]{graphicx}
\usepackage{hyperref}

\title{HistoSpeckle-Net: Mutual Information-Guided Deep Learning for high-fidelity reconstruction of complex OrganAMNIST images via perturbed Multimode Fibers}

\author[1]{Jawaria Maqbool}
\author[1]{M.~Imran Cheema\thanks{Corresponding author: imran.cheema@lums.edu.pk}}
\affil[1]{Department of Electrical Engineering, Syed Babar Ali School of Science and Engineering,\\
Lahore University of Management Sciences, Lahore, Pakistan}

\begin{document}

\maketitle

\begin{abstract}
Existing deep learning methods in multimode fiber (MMF) imaging often focus on simpler datasets, limiting their applicability to complex, real-world imaging tasks. These models are typically data-intensive, a challenge that becomes more pronounced when dealing with diverse and complex images. In this work, we propose HistoSpeckle-Net, a deep learning architecture designed to reconstruct structurally rich medical images from MMF speckles. To build a clinically relevant dataset, we develop an optical setup that couples laser light through a spatial light modulator (SLM) into an MMF, capturing output speckle patterns corresponding to input OrganAMNIST images. Unlike previous MMF imaging approaches, which have not considered the underlying statistics of speckles and reconstructed images, we introduce a distribution-aware learning strategy. We employ a histogram-based mutual information loss to enhance model robustness and reduce reliance on large datasets. Our model includes a histogram computation unit that estimates smooth marginal and joint histograms for calculating mutual information loss. It also incorporates a unique Three-Scale Feature Refinement Module, which leads to multiscale Structural Similarity Index Measure (SSIM) loss computation. Together, these two loss functions enhance both the structural fidelity and statistical alignment of the reconstructed images. Our experiments on the complex OrganAMNIST dataset demonstrate that HistoSpeckle-Net achieves higher fidelity than baseline models such as U-Net and Pix2Pix. It gives superior performance even with limited training samples and across varying fiber bending conditions. By effectively reconstructing complex anatomical features with reduced data and under fiber perturbations, HistoSpeckle-Net brings MMF imaging closer to practical deployment in real-world clinical environments.
\end{abstract}

\section{Introduction}
\label{introduction}
Recent advancements in endoscopic technology have improved medical diagnostics by enabling minimally invasive visualization of internal organs and tissues. Among these innovations, multimode fiber~(MMF) imaging has emerged as a promising approach due to MMF's thin structure and potential for high-resolution image transmission \cite{hadley1965gastro,gu2014fibre,perperidis2020image}. However, a significant challenge in their application is reconstructing images from the complex speckle patterns generated at the distal end of the fiber, which result from modal dispersion and interference. Traditional methods, such as optical phase conjugation and transmission matrix approaches, have been used to address the aforementioned challenge. However, they require precise interferometric setups for phase measurements and are highly sensitive to external perturbations \cite{papadopoulos2012focusing,popoff2010measuring,akbulut2013measurements,dremeau2015reference}. Deep learning offers robust solutions to overcome these limitations and provide reliable reconstruction of images from speckle patterns \cite{rahmani2018multimode}.

Ongoing progress in deep learning has improved the performance of MMF imaging systems. Prior works have explored various neural network architectures for this purpose \cite{fan2019deep,caramazza2019transmission,abdulaziz2023robust,li2024feature,kremp2023neural}, including a single hidden layer dense neural network \cite{zhu2021image}, the U-Net model \cite{zhu2023deep}, an attention-based U-Net architecture \cite{song2022deep}, and a conditional generative adversarial network \cite{maqbool2024towards}. For color imaging through MMF, researchers have introduced SpeckleColorNet \cite{zhang2024wide}, and more recently, a multi-speckle illumination type inverse transmission matrix-Unet method is proposed for illumination and imaging through a single multimode fiber \cite{feng2025high}. These works have primarily focused on relatively simpler datasets such as MNIST and Fashion-MNIST to illustrate the basic functionality of the proposed models. While some studies have expanded to natural scenes and peripheral blood cells images, the transition to medical imaging applications presents unique challenges due to the complexity and fine detail present in medical images. This complexity is particularly evident when considering the diverse features of different organs and tissues that must be accurately reconstructed for reliable diagnostic purposes \cite{fallahpoor2024deep}. For MMF imaging systems to be effectively integrated into practical medical applications like endoscopy, training these systems on datasets that closely mirror real-world use cases is essential. Therefore,  in the current work, we utilize the diverse and multiclass dataset, OrganAMNIST, which consists of CT axial images of various organs \cite{yang2023medmnist}.

These complex images are transformed into speckle patterns by multimode fibers, and while the resulting intensity distributions may look random, they follow the beta distribution. It is due to the finite input power and the bounded nature of scattered wave amplitudes within the fiber core \cite{goodman2007speckle}. This statistical behavior underscores the relevance of histogram analysis when working with speckle-based imaging. However, a key challenge arises: the reconstructed images from these speckles and the original OrganAMNIST labels do not inherently follow a beta distribution. Moreover, their discrete histograms and fitted beta distributions are not differentiable, making them unsuitable for gradient-based learning frameworks. To address this, we propose a method for computing smooth and differentiable approximations of image histograms that preserve the structure of the original distributions.

In addition, the distribution of the speckle patterns and the images reconstructed from them can vary with changes in physical factors such as temperature, fiber length, and fiber bending. These underlying statistics motivate the use of a distribution-based loss function. Mutual information loss, derived from information theory, is well-suited in this case as it measures the amount of shared information between two signals and captures complex statistical dependencies. It is also inherently robust to lighting variations and nonlinear transformations \cite{viola1997alignment}. This work introduces a novel HistoSpeckle-Net that combines multiscale structural similarity~(MS-SSIM) loss and a mutual information loss~(MI) calculated through differentiable smooth histograms of the reconstructed and ground truth images. While SSIM loss encourages preservation of texture and structure, the histogram-based mutual information loss ensures that the statistical distributions of generated images closely match those of the targets. By maximizing the shared entropy between distributions, MI captures higher-order statistical relationships. This distribution-level alignment helps the model capture the physical constraints inherent in speckle formation, improving reconstruction fidelity, especially for structurally complex datasets like OrganMNIST.

Our proposed HistoSpeckle-Net incorporates a unique Three-Scale Feature Refinement Module~(TFRM), connected to the final four decoding layers of the generator. This targeted approach focuses on enhancing high-level semantic features while reducing parameter count.  This targeted refinement strategy contrasts methods that broadly apply feature enhancement, often leading to computational inefficiencies \cite{chen2020laplacian,xing2021gated}. It also facilitates the calculation of Structural Similarity Index Measure~(SSIM) loss at three resolution levels, contributing to improved reconstruction accuracy. The network also includes a histogram computation unit~(HCU) at the generator's output, which calculates differentiable marginal and joint histograms for the generated images and their corresponding ground truth labels. These histograms are then used to compute the mutual information loss. Histogram-based losses and network architectures have been previously explored in tasks such as color transfer \cite{afifi2021histogan,avi2023differentiable}, facial recognition \cite{sadeghi2022histnet}, and image-to-image translation \cite{peng2023histogram}. Histogram matching has recently been employed for endo-microscopy \cite{chung2025histogram}. However, mutual information loss, originally introduced in \cite{viola1997alignment} and more recently adopted for color transfer in \cite{avi2023differentiable}, remains unexplored in the context of MMF imaging.

This work demonstrates that our deep learning model, equipped with a uniquely crafted loss function, achieves high-fidelity reconstruction of complex OrganMNIST images. The architecture integrates mutual information~(MI) loss with multiscale structural similarity~(MS-SSIM) loss, collectively promoting both structural and statistical consistency. By enforcing distributional alignment between the reconstructed and ground truth data, our method facilitates the accurate recovery of fine structural details. This holds true not only under ideal, static MMF imaging conditions but also in the presence of fiber perturbations and data-scarce regimes. Comparative evaluations of our model with widely adopted deep learning architectures, including U-Net and the Pix2Pix conditional generative adversarial network, reveal clear performance gains. Our demonstrated HistoSpeckle-Net achieves an average SSIM of 0.7240, outperforming U-Net~(0.6416) and Pix2Pix~(0.5822). Under perturbed fiber conditions, HistoSpeckle-Net shows an SSIM above 0.64 across all bending positions, demonstrating robust performance. In contrast, U-Net and Pix2Pix show a noticeable decline in image quality under the same conditions.
\section{Experimental Setup}
\begin{figure*}[htbp]
\centering\includegraphics[width=5.9in]{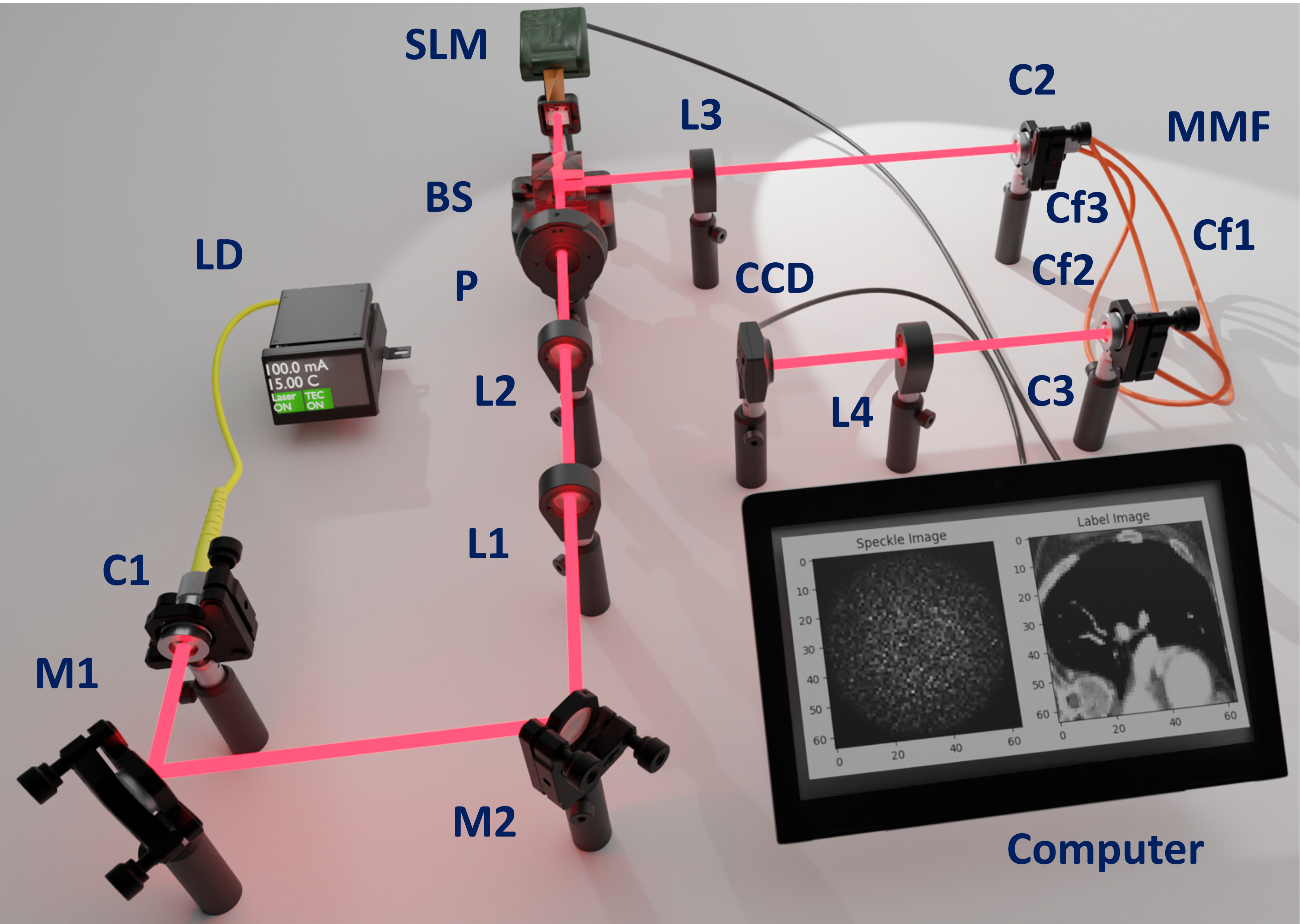}
\caption{Experimental setup for data collection corresponding to three different multimode fiber (MMF) configurations. LD: Laser diode, C: Collimator, L: Lens, M: Mirror, BS: Beam splitter, P: Linear polarizer, SLM: Spatial light modulator, CCD: Camera, MMF:  Multimode fiber, Cf: Fiber configuration.}
\label{setup}
\end{figure*}
\begin{figure*}[htbp]
\centering\includegraphics[width=5.9in]{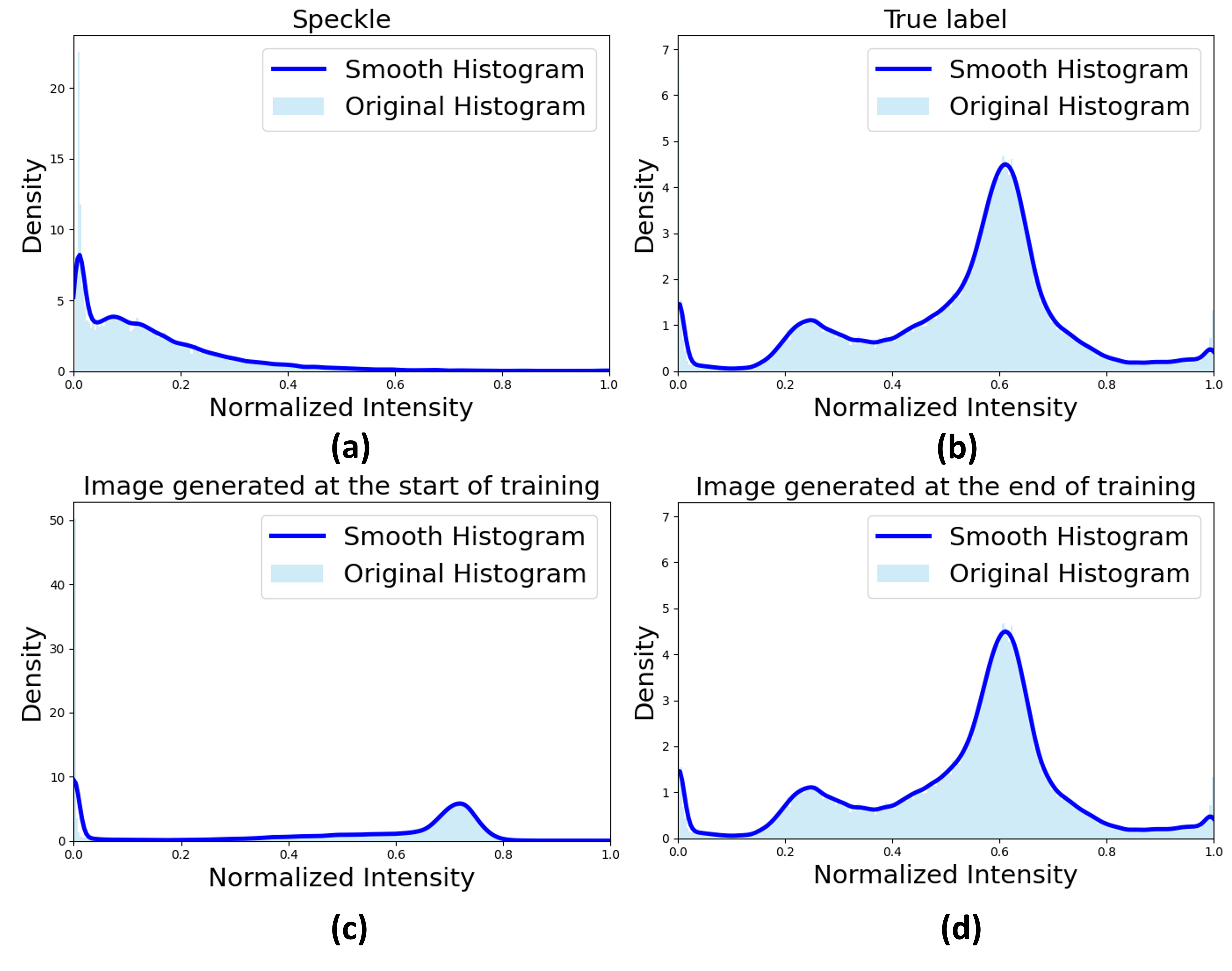}
\caption{Comparison of normalized intensity histograms for (a) speckle input,(b) true label,(c) initially generated image, and (d) final generated image. The original histograms (shaded area) and smooth differentiable histograms (blue line) are shown.  Histogram alignment with ground truth (b) improves from the (c) initial to the (d) final generated images, indicating learning progression.}
\label{histo}
\end{figure*}

\begin{figure*}[htbp]
\centering\includegraphics[width=7.0in]{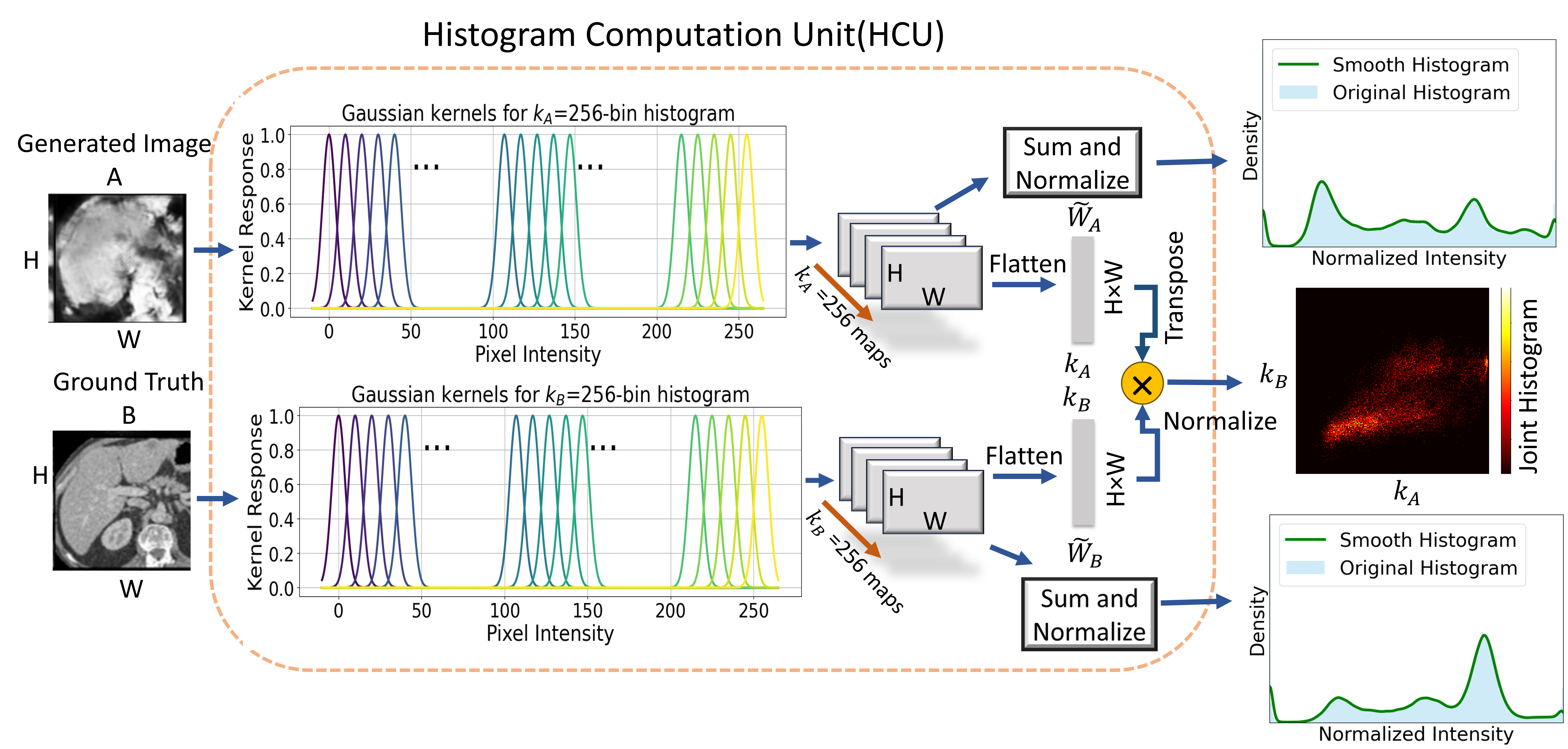}
\caption{An overview of computing smooth marginal histograms and the joint histogram for both the generated and ground truth images using a histogram computation unit. Each image is passed through a bank of 256 Gaussian kernels to generate kernel response maps. The kernel responses are then summed and normalized to produce smooth histograms (green curves), which closely match the original discrete histograms (light blue bars). These maps are also flattened into matrices \( \tilde{W}_A \) and \( \tilde{W}_B \), representing smooth assignments to histogram bins. The flattened matrices are multiplied and normalized to produce the final joint histogram.}
\label{hcu}
\end{figure*}
The schematic of the experimental setup is presented in Fig.~\ref{setup}. A laser diode with a wavelength 633~nm (Eagleyard GC-02940), driven by a Thorlabs CLD1015 controller, serves as the light source. The emitted laser beam is directed using mirrors and subsequently collimated by a telescopic arrangement consisting of two lenses with focal lengths of 500~mm and 100~mm, respectively. A polarizer is positioned after the telescope to ensure polarization alignment with the HOLOEYE Pluto 2.0 spatial light modulator~(SLM). The polarized beam is then guided to a 50/50 beam splitter~(BS), where half of the beam is transmitted towards the SLM while the other half is blocked. The SLM reflects the phase-modulated beam, which then passes back through the BS. A third lens focuses this modulated light onto a second collimator, which couples the beam into a multimode fiber~(MMF). The MMF used in the setup has a core diameter of 400~$\mu$m, a numerical aperture (NA) of 0.22, and a length of 1~m.

We use the OrganAMNIST dataset, which consists of 58,830 grayscale axial view images of various organs, including the right femoral head, heart, left femoral head, bladder, left kidney, right kidney, liver, left lung, right lung, spleen, and pancreas. These are encoded onto the laser beam via the SLM. The encoded image information is scrambled into complex speckle patterns as the light travels through the MMF. These output speckle patterns emerge from the distal end of the fiber, are imaged through another lens onto a Thorlabs DCC1545M CMOS camera. The resulting speckle images are saved on a computer to form the output dataset. To simulate realistic endoscopic conditions where the fiber may bend or shift, the experiment is repeated for three distinct MMF configurations. As a result, three separate datasets of OrganMNIST-based speckle images are generated, each corresponding to a different MMF position.
\section{Histogram computation unit}

We examine their intensity histograms to leverage the statistical characteristics of speckle patterns captured at a multimode fiber's output. Traditional discrete histograms and their fitted distributions are non-differentiable and cannot be directly used in gradient-based training. To address this limitation, we compute smooth, differentiable histograms for raw speckles, the initially generated image, the true label, and the final output after training using a Gaussian kernel \cite{avi2023differentiable}. These smoothed histograms, plotted in blue in Fig.~\ref{histo}, follow the shape of the original discrete histograms. Notably, while the initially generated images differ significantly from the true label in terms of intensity distribution, the final generated outputs closely match the ground-truth histogram, indicating effective learning during training.

The smooth histograms are calculated using a histogram computation unit whose architecture is shown in Fig.~\ref{hcu}. Let $I_A(p)$ represent the intensity of image $A$ at pixel location $p$. We use $k_A$ to denote the number of bins in the histogram for image $A$, and $b_i$ to represent the center value of the $i$-th bin. The weight assigned to bin $i$ for pixel $p$ in image $A$, which we denote as $W_A(p, i)$, is calculated as:
\begin{equation}
W_A(p, i) = \exp\left(-\frac{1}{2} \left(\frac{I_A(p) - b_i}{\sigma}\right)^2\right)
\end{equation}
where $\sigma$ controls the width of the Gaussian kernel, determining how much a pixel's influence is spread to neighboring bins. To ensure that each pixel contributes a total weight of one, we normalize these weights:
\begin{equation}
\tilde{W}_A(p, i) = \frac{W_A(p, i)}{\sum_{i'=0}^{k_A-1} W_A(p, i')}
\end{equation}
The smooth histogram value for bin $i$, denoted as $\tilde{H}_A(i)$, is then the sum of these normalized weights across all pixels:
\begin{equation}
\tilde{H}_A(i) = \sum_{p \in [0, 1]} \tilde{W}_A(p, i)
\end{equation}
Finally, we normalize the histogram to obtain the  marginal probability distribution, $\tilde{P}_A(i)$:
\begin{equation}
\tilde{P}_A(i) = \frac{\tilde{H}_A(i)}{\sum_{i'=0}^{k_A-1} \tilde{H}_A(i')}
\label{eq:eq4}
\end{equation}
This smooth histogram, $\tilde{P}_A(i)$, represents the probability of observing intensity value $i$ in image $A$, but in a soft manner. We use $k_A = 256$ histogram bins with linearly spaced centers over the intensity range $[0, 1]$, such that $i = 0, 1, \ldots, 255$. In our implementation, we set $\sigma = 0.01$, which achieves an effective trade-off between bin separation and smooth gradient propagation.

In addition to marginal histogram calculation, our HCU also computes the statistical relationship~(joint histogram) of two images which is necessary for the calculation of mutual information loss. For two images $A$ and $B$, the joint histogram captures the frequency with which each possible pair of intensity values appears at corresponding pixel positions across both images. As shown in Fig.~\ref{hcu},  $\mathbf{\tilde{W}}_A$ be a matrix where each element represents the normalized weight $\tilde{W}_A(p, i)$ of pixel $p$ assigned to bin $i$ in image $A$ and is obtained after flattening $k_A$ activation maps.  We define $\mathbf{\tilde{W}}_B$ similarly for image B.  The smooth joint histogram, $\tilde{H}_{AB}(i, j)$, which indicates the frequency with which intensity bin $i$ in image $A$ co-occurs with intensity bin $j$ in image $B$, is then computed as:
\begin{equation}
\tilde{H}_{AB}(i, j) = \sum_{p \in [0, 1]}  \tilde{W}_A(p, i) \cdot \tilde{W}_B(p, j)
\end{equation}
This operation is efficiently implemented using a matrix multiplication:
\begin{equation}
\tilde{H}_{AB}(i, j) = \tilde{W}_A  \tilde{W}_B^T
\end{equation}
The resulting joint histogram is then normalized to obtain the soft joint probability distribution, $\tilde{P}_{AB}(i, j)$:
\begin{equation}
\tilde{P}_{AB}(i, j) = \frac{\tilde{H}_{AB}(i, j)}{\sum_{i'=0}^{k_A-1} \sum_{j'=0}^{k_B-1} \tilde{H}_{AB}(i', j') }
\label{eq:eq7}
\end{equation}

\section{Deep Learning Framework}
\begin{figure*}[htbp]
\centering\includegraphics[width=7.0in]{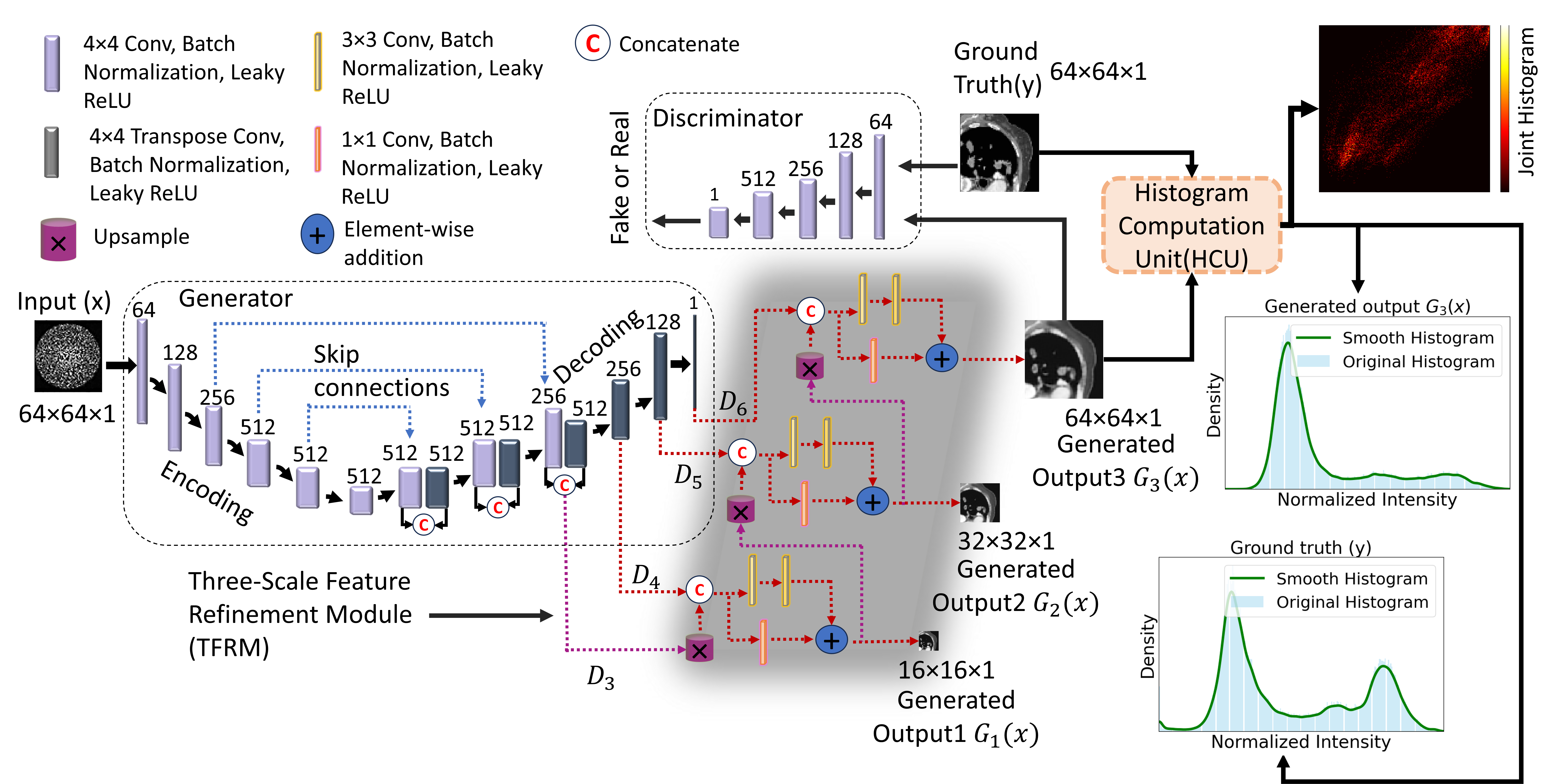}
\caption{The complete architecture of HistoSpeckle-Net}
\label{network}
\end{figure*}
Our proposed HistoSpeckle-Net framework is illustrated in Fig.~\ref{network}. The network takes speckle patterns as an input to a U-Net-type generator consisting of an encoder-decoder structure. The decoding stage is enhanced with a Three-Scale Feature Refinement Module, connected to the last four decoding layers. This module progressively enhances features and produces three different scale outputs. The highest resolution output $G_3(x)$, which is also considered the final predicted output, is passed to a discriminator along with the ground-truth image to determine whether the image is real or generated. The discriminator is a PatchGAN model, which classifies patches of the input image rather than the entire image, improving image detail and structure. In addition, marginal and joint histograms for generated $G_3(x)$ and ground truth $y$ images are computed using a Histogram Computation Unit (HCU). These are fed into a mutual information loss function to ensure statistical consistency between generated and real images.
\subsection{Three-Scale Feature Refinement Module (TFRM)}
Our network features a specialized Three-scale Feature Refinement Module (TFRM) that attaches to the final four decoding layers $(D_3$, $D_4$, $D_5$, and $D_6)$, producing outputs at three different scales $(G_1(x)$, $G_2(x)$, and $G_3(x))$. Unlike previous approaches that apply progressive refinement across all decoding layers \cite{chen2020laplacian,xing2021gated}, our TFRM significantly reduces computational cost while maintaining accuracy. TFRM progressively incorporates information from three different frequency bands. The low-scale output captures low-frequency signals, while higher-resolution outputs retain higher-frequency information. This progressive addition and refinement process facilitates our predicted images in achieving pixel-wise accuracy and preserving fine details and structures.

The enhanced features of the previous scale, $G_{m-1}(x)$, are first upsampled~($\widetilde{U}$) using bilinear interpolation and concatenated~($||$) with decoding features of the current scale,$D_l$.  These features are further refined by a residual block consisting of two consecutive 3×3 convolutional layers in the main path. In parallel, a shortcut path includes a 1×1 convolution to match the dimensionality of the main path. This residual connection ensures effective gradient flow and helps the network to learn low and high-frequency details, contributing to accurate reconstruction at each scale. The outputs of both paths are then combined through element-wise addition to produce the final refined feature map $G_{m}(x)$ :
\begin{align}
G_m(x) &= \text{Conv}_{1\times1} \left( \widetilde{U}(G_{m-1}(x)) \,\|\, D_l \right) \nonumber\\
&\quad + \text{Conv}_{3\times3} \left( \text{Conv}_{3\times3} \left( \widetilde{U}(G_{m-1}(x)) \,\|\, D_l \right) \right)
\end{align}
where $m=2,3$ and $l=5,6$. Initially, at the lowest resolution level, where no refined features are available, the process begins by upsampling the decoding features $D_3$ and concatenating them with the next higher resolution decoding features $D_4$. This combined representation is refined to produce the initial enhanced map $G_1(x)$ as follows:
\begin{align}
G_1(x) &= \text{Conv}_{1\times1} \left( \widetilde{U}(D_3) \,\|\, D_4 \right) \nonumber\\
&\quad + \text{Conv}_{3\times3} \left( \text{Conv}_{3\times3} \left( \widetilde{U}(D_3) \,\|\, D_4 \right) \right)
\end{align}
It is important to note that the highest-resolution refined feature map, $G_3(x)$, becomes the final output prediction.
\subsection{Loss function}
The quality of image reconstruction in our framework is significantly influenced by the design of the loss function. Therefore, we propose a carefully constructed composite loss that combines adversarial loss, mutual information loss, and multiscale SSIM loss for training the generator. For the discriminator, we adopt the standard binary cross-entropy (BCE) loss. The adversarial component encourages the generator to produce realistic images. We use binary cross-entropy loss as the adversarial loss for the generator, defined as:
\begin{equation}
\mathcal{L}_{\text{adv}} = \text{BCE}(D(G_3(x), x), 1)
\end{equation}
where $G_3(x)$ denotes the final output of the generator and $D()$ represents the discriminator’s output .

Mutual information (MI) is the key component of our loss function. It measures the amount of information shared between two variables. In our case, it quantifies the dependency between the reconstructed image $G_3(x)$ and the ground truth image $y$ and is defined as:

\begin{equation}
I(y, G_3(x)) = H(y) - H(y|G_3(x))
\end{equation}

where $H(y)$ is the entropy of the source image and $H(y|G_3(x))$ is the conditional entropy.  Maximizing mutual information helps in preserving global structure and intensity distribution. The entropy of the ground truth image, $H(y)$, is assumed to be constant. Therefore, instead of directly maximizing $I(y, G_3(x))$, we minimize the conditional entropy $H(y|G_3(x))$. A lower conditional entropy value indicates that the output image provides more information about the source image, implying greater similarity \cite{avi2023differentiable}. The conditional entropy is calculated using the joint~($\tilde{P}_{y G_3(x)}(i, j) $) and marginal probability distributions~($\tilde{P}_{G_3(x)}(j) $) from Eqs.~\eqref{eq:eq4} and \eqref{eq:eq7} and is given by:
\begin{figure*}[htbp]
\centering\includegraphics[width=6.3in]{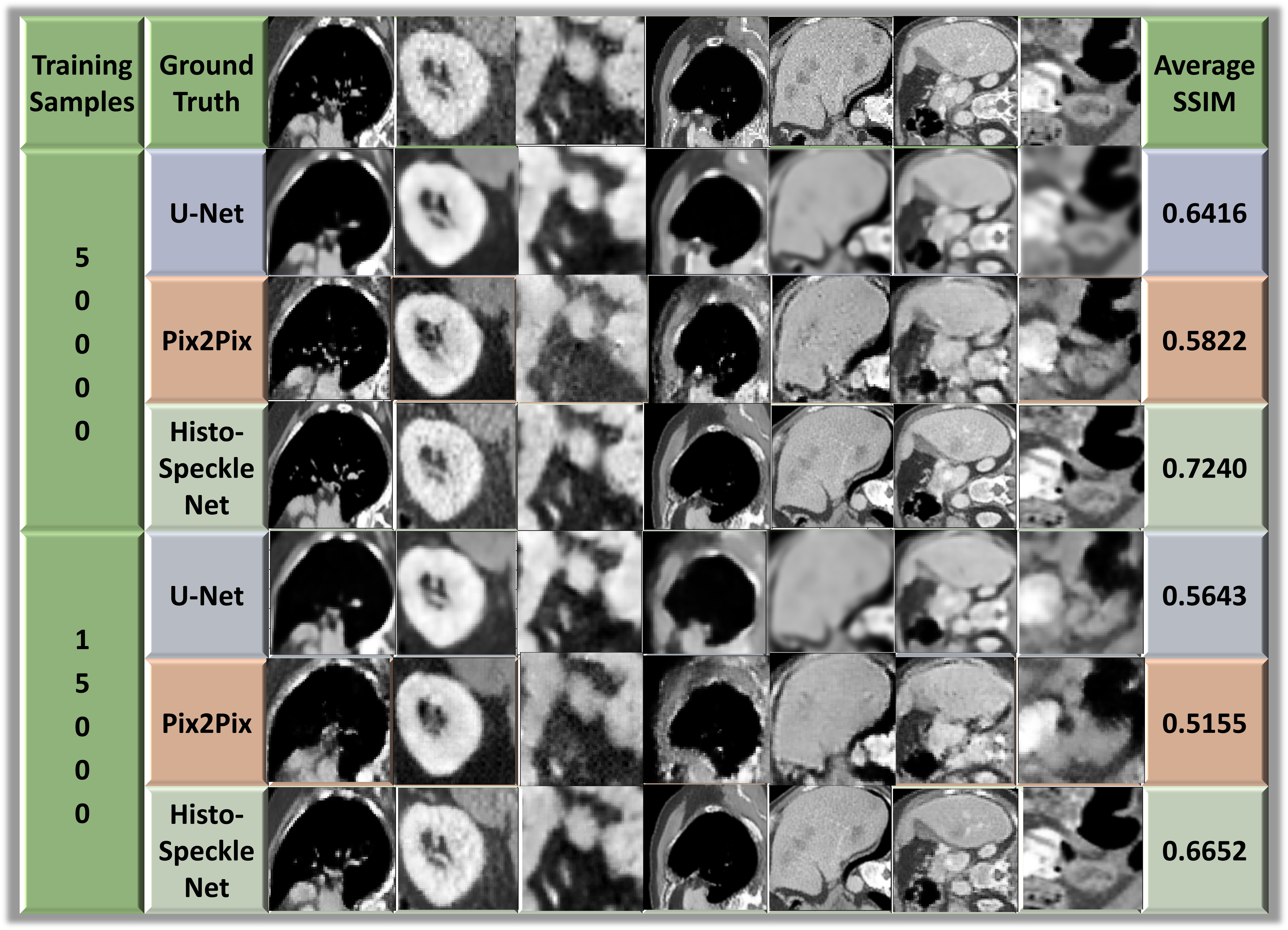}
\caption{Reconstruction results for a fixed fiber and for different numbers of training samples.}
\label{results1}
\end{figure*}

\begin{align}
\mathcal{L}_{\text{MI}}
&= H(y|G_3(x)) \nonumber\\
&= - \sum_{i=0}^{k_y - 1} \sum_{j=0}^{k_{G_3(x)} - 1}
\tilde{P}_{y G_3(x)}(i, j) \cdot
\log_2\left( \frac{\tilde{P}_{y G_3(x)}(i, j)}{\tilde{P}_{G_3(x)}(j)} \right)
\end{align}

Minimizing this conditional entropy, $H(y|G_3(x))$, effectively maximizes the mutual information between the source image, $y$, and the output image, $G_3(x)$, thus indicating a greater degree of similarity.

To preserve accurate structural details, we incorporate the multiscale structural similarity (MS-SSIM) loss, which evaluates the luminance, contrast, and structure at multiple resolutions. We apply this loss across all three outputs $G_m(x)$ ($m = 1, 2, 3$) from our feature refinement module. To accurately compute the MS-SSIM loss, we first upsample the lower-resolution outputs $G_1(x)$ and $G_2(x)$ to match the spatial dimensions of the ground truth image $y$. By enforcing structural alignment at each scale, the MS-SSIM loss guides the network to preserve fine details and contextual structure, which may be lost with a single-scale SSIM approach.
\begin{equation}
\mathcal{L}_{\text{SSIM}} = \sum_{m=1}^{3} \text{MS-SSIM}(G_m(x), y)
\end{equation}
The overall generator loss function combines the three components discussed above:
\begin{equation}
\mathcal{L}_{\text{Gen}} = \mathcal{L}_{\text{adv}} + \lambda_1 \mathcal{L}_{\text{MI}} + \lambda_2 \mathcal{L}_{\text{SSIM}}
\end{equation}
where $\lambda_1$ and $\lambda_2$ are weighting coefficients that balance the contribution of mutual information and structural similarity terms.
For the discriminator, we use the binary cross-entropy loss to differentiate between real and fake~(generated) images:
\begin{equation}
\mathcal{L}_{\text{Dis}} = \frac{1}{2} \, \text{BCE}(D(y, x), 1) + \frac{1}{2} \, \text{BCE}(D(G_3(x), x), 0)
\end{equation}
The total loss of the model is the sum of the discriminator and generator losses.
\begin{equation}
\mathcal{L}_{\text{HistoSpeckle-Net}} = \mathcal{L}_{\text{Dis}} + \mathcal{L}_{\text{Gen}}
\end{equation}

\section{Reconstruction results }
\begin{figure*}[htbp]
\centering\includegraphics[width=6.3in]{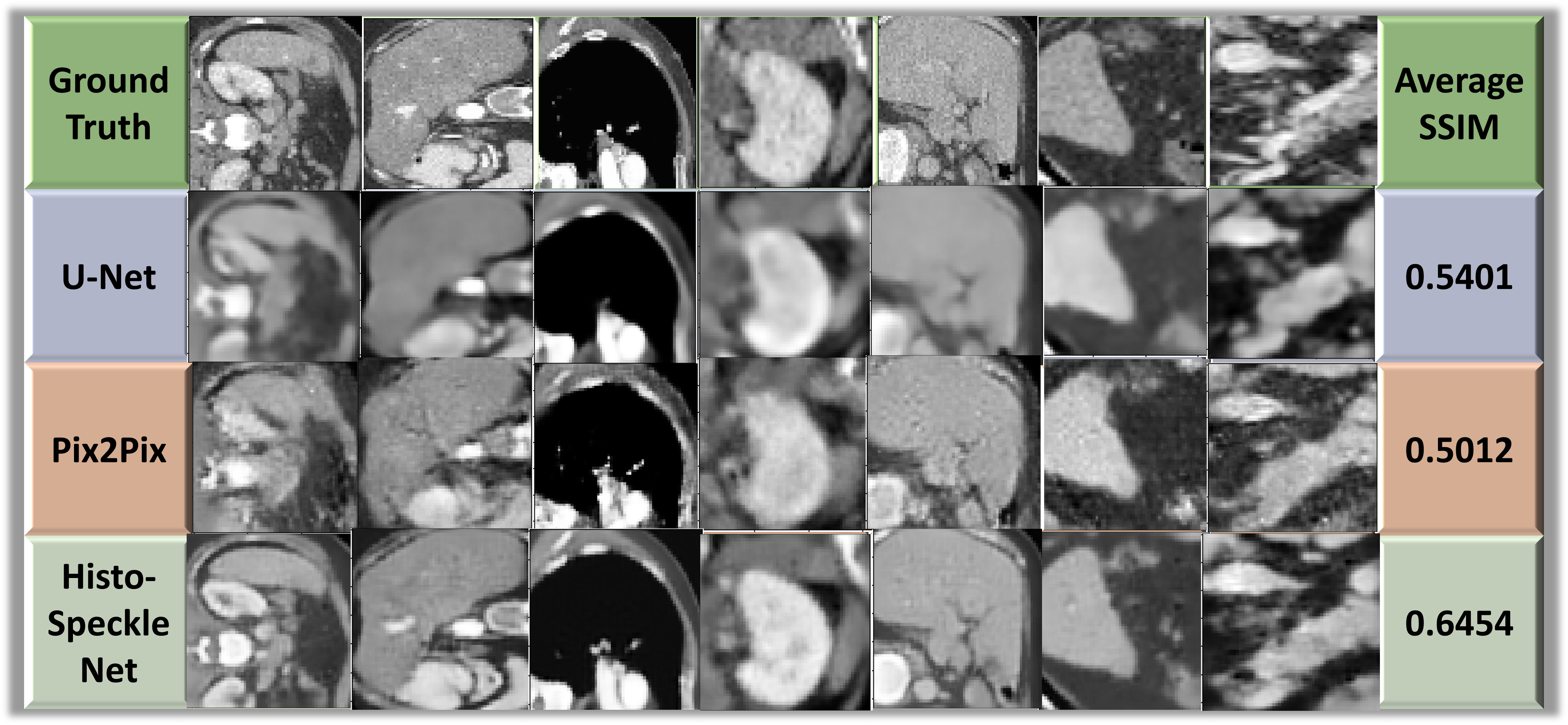}
\caption{Reconstruction results for a perturbed fiber}
\label{results2}
\end{figure*}
To evaluate the performance of our proposed model, we perform image reconstruction experiments using OrganMNIST images. These experiments assess our model's ability to maintain high fidelity under both ideal and challenging conditions, specifically when imaging through a fixed multimode fiber and under conditions where the fiber was subject to perturbations. Furthermore, we investigate the model's capacity to generate accurate images even when trained on a limited dataset. Using our experimental setup, we collect three distinct datasets, each consisting of 58,830 speckle–organ image pairs, corresponding to three different configurations of the multimode fiber. To ensure robust training and evaluation, we partition each dataset into training (50,000 samples), validation (2,947 samples), and testing (5,883 samples) subsets.

We begin by training and evaluating each model independently for every fiber position. In this initial set of experiments, we train Histo-Speckle Net, a standard U-Net, and a conditional GAN (pix2pix) model using the full training set of 50,000 samples. Their reconstruction performance is then assessed on the test set, with the Structural Similarity Index (SSIM) serving as the primary evaluation metric. Histo-Speckle Net outperforms the baseline models, achieving an average SSIM of 0.7240, compared to 0.6416 for U-Net and 0.5822 for pix2pix on unseen test images. Visual inspection of the reconstructed images reveals further insights. While the baseline U-Net model produces a better SSIM score than pix2pix, the reconstructed details appear smoothed out. The pix2pix model, on the other hand, generates images with more pronounced details, but the reconstruction accuracy of certain features is limited. In contrast, Histo-Speckle Net generates high-fidelity reconstructions, preserving fine details without introducing excessive smoothing as given in Fig.~\ref{results1}.

To assess the models' performance with limited data, we train them on a reduced training set consisting of only 15,000 randomly selected samples (30\% of the original training set). The results in Fig.~\ref{results1} demonstrate that Histo-Speckle Net maintains a superior average SSIM of 0.6652 as compared to both U-Net~(0.5643) and pix2pix~(0.5155). Notably, Histo-Speckle Net exhibits reasonable image fidelity even with the reduced dataset, highlighting its robustness and suitability for applications where data availability is limited.

Finally, to simulate the effects of perturbations or bending in multimode fiber imaging, we create a combined dataset by randomly sampling only 12,000 images from each of the three fiber configurations. We train our model on this combined dataset and evaluate its performance on the original and separated test datasets~(5,883 samples per fiber position). The average SSIM for Histo-Speckle Net remains above 0.64 for each test data of the separate fiber configuration. We compare the performance of our model under these perturbation conditions with that of U-Net and pix2pix, which exhibit lower average SSIM scores as shown in Fig.~\ref{results2}. These results highlight the superior performance of our model in handling the challenges posed by perturbations and bending in multimode fiber imaging.

\section{Conclusion}
Our work demonstrates that incorporating distribution-aware learning with mutual information loss, alongside multiscale structural similarity losses, significantly enhances image reconstruction fidelity in MMF imaging. HistoSpeckle-Net's ability to preserve fine structural details of complex OrganAMNIST images, even with limited training data and under fiber perturbations, marks a significant advancement in this field. Our results highlight the effectiveness of combining histogram-based loss functions, grounded in the physical and statistical behavior of MMF speckles, with architectural enhancements such as the Three-Scale Feature Refinement Module. This integrated approach improves the robustness of deep learning models in challenging imaging scenarios and moves MMF imaging closer to clinical deployment, especially in settings where the acquisition of large annotated datasets is impractical. However, some open questions remain. For example, how well the model generalizes to other types of medical images or different fiber geometries is yet to be explored. While our method shows improved performance under typical perturbations, understanding its limitations under extreme conditions can lead to future refinements. Looking ahead, HistoSpeckle-Net can be used for other scattering-media imaging (e.g., turbid fluids, biological tissues) and color imaging through MMFs. Moreover, the architectural strategies proposed here may benefit other domains requiring high-fidelity image-to-image translation, such as low-light microscopy, remote sensing, image super-resolution, medical imaging, and segmentation.

\medskip
\textbf{Acknowledgements} \par 
This work is supported by Syed Babar Ali Research Award (SBARA)(GRA-0085) for the year 2024-2025.

\medskip

\bibliographystyle{unsrt}
\bibliography{synthetic}

@article{hadley1965gastro,
  title={The gastro-camera.},
  author={Hadley, GD},
  journal={British Medical Journal},
  volume={2},
  number={5472},
  pages={1209},
  year={1965},
  publisher={BMJ Publishing Group}
}

@article{gu2014fibre,
  title={Fibre-optical microendoscopy},
  author={Gu, Min and Bao, Hongchun and Kang, Hong},
  journal={Journal of microscopy},
  volume={254},
  number={1},
  pages={13--18},
  year={2014},
  publisher={Wiley Online Library}
}

@article{perperidis2020image,
  title={Image computing for fibre-bundle endomicroscopy: A review},
  author={Perperidis, Antonios and Dhaliwal, Kevin and McLaughlin, Stephen and Vercauteren, Tom},
  journal={Medical image analysis},
  volume={62},
  pages={101620},
  year={2020},
  publisher={Elsevier}
}

@article{papadopoulos2012focusing,
  title={Focusing and scanning light through a multimode optical fiber using digital phase conjugation},
  author={Papadopoulos, Ioannis N and Farahi, Salma and Moser, Christophe and Psaltis, Demetri},
  journal={Optics express},
  volume={20},
  number={10},
  pages={10583--10590},
  year={2012},
  publisher={Optica Publishing Group}
}

@article{popoff2010measuring,
  title={Measuring the transmission matrix in optics: an approach to the study and control of light propagation in disordered media},
  author={Popoff, S{\'e}bastien M and Lerosey, Geoffroy and Carminati, R{\'e}mi and Fink, Mathias and Boccara, Albert Claude and Gigan, Sylvain},
  journal={Physical review letters},
  volume={104},
  number={10},
  pages={100601},
  year={2010},
  publisher={APS}
}

@inproceedings{akbulut2013measurements,
  title={Measurements on the optical transmission matrices of strongly scattering nanowire layers},
  author={Akbulut, Duygu and Strudley, Tom and Bertolotti, Jacopo and Zehender, Tilman and Bakkers, Erik PAM and Lagendijk, Ad and Vos, Willem L and Muskens, Otto L and Mosk, Allard P},
  booktitle={International Quantum Electronics Conference},
  pages={IH\_P\_19},
  year={2013},
  organization={Optica Publishing Group}
}

@article{dremeau2015reference,
  title={Reference-less measurement of the transmission matrix of a highly scattering material using a DMD and phase retrieval techniques},
  author={Dr{\'e}meau, Ang{\'e}lique and Liutkus, Antoine and Martina, David and Katz, Ori and Sch{\"u}lke, Christophe and Krzakala, Florent and Gigan, Sylvain and Daudet, Laurent},
  journal={Optics express},
  volume={23},
  number={9},
  pages={11898--11911},
  year={2015},
  publisher={Optica Publishing Group}
}

@article{rahmani2018multimode,
  title={Multimode optical fiber transmission with a deep learning network},
  author={Rahmani, Babak and Loterie, Damien and Konstantinou, Georgia and Psaltis, Demetri and Moser, Christophe},
  journal={Light: science \& applications},
  volume={7},
  number={1},
  pages={69},
  year={2018},
  publisher={Nature Publishing Group UK London}
}

@article{fan2019deep,
  title={Deep learning the high variability and randomness inside multimode fibers},
  author={Fan, Pengfei and Zhao, Tianrui and Su, Lei},
  journal={Optics express},
  volume={27},
  number={15},
  pages={20241--20258},
  year={2019},
  publisher={Optica Publishing Group}
}

@article{zhu2021image,
  title={Image reconstruction through a multimode fiber with a simple neural network architecture},
  author={Zhu, Changyan and Chan, Eng Aik and Wang, You and Peng, Weina and Guo, Ruixiang and Zhang, Baile and Soci, Cesare and Chong, Yidong},
  journal={Scientific reports},
  volume={11},
  number={1},
  pages={896},
  year={2021},
  publisher={Nature Publishing Group UK London}
}

@article{song2022deep,
  title={Deep learning image transmission through a multimode fiber based on a small training dataset},
  author={Song, Binbin and Jin, Chang and Wu, Jixuan and Lin, Wei and Liu, Bo and Huang, Wei and Chen, Shengyong},
  journal={Optics express},
  volume={30},
  number={4},
  pages={5657--5672},
  year={2022},
  publisher={Optica Publishing Group}
}

@article{zhu2023deep,
  title={Deep learning-based multimode fiber imaging in multispectral and multipolarimetric channels},
  author={Zhu, Run-ze and Feng, Hao-gong and Xu, Fei},
  journal={Optics and Lasers in Engineering},
  volume={161},
  pages={107386},
  year={2023},
  publisher={Elsevier}
}

@inproceedings{kremp2023neural,
  title={Neural-network-based multimode fiber imaging and position sensing under thermal perturbations},
  author={Kremp, Tristan and Bagley, Nicholas and Lamb, Erin S and Westbrook, Paul S and DiGiovanni, David J},
  booktitle={Adaptive Optics and Wavefront Control for Biological Systems IX},
  volume={12388},
  pages={35--48},
  year={2023},
  organization={SPIE}
}

@article{abdulaziz2023robust,
  title={Robust real-time imaging through flexible multimode fibers},
  author={Abdulaziz, Abdullah and Mekhail, Simon Peter and Altmann, Yoann and Padgett, Miles J and McLaughlin, Stephen},
  journal={Scientific Reports},
  volume={13},
  number={1},
  pages={11371},
  year={2023},
  publisher={Nature Publishing Group UK London}
}

@article{feng2025high,
  title={High-fidelity image reconstruction in multimode fiber imaging through the MITM-Unet framework},
  author={Feng, Zefeng and Yue, Zengqi and Zhou, Wei and Xu, Baoteng and Liu, Jialin and Sun, Jiawei and Xiong, Daxi and Yang, Xibin},
  journal={Optics Express},
  volume={33},
  number={3},
  pages={5866--5876},
  year={2025},
  publisher={Optica Publishing Group}
}

@article{zhang2024wide,
  title={Wide-field color imaging through multimode fiber with single wavelength illumination: plug-and-play approach},
  author={Zhang, Hailong and Wang, Lele and Xiao, Qirong and Ma, Jianshe and Zhao, Yi and Gong, Mali},
  journal={Optics Express},
  volume={32},
  number={4},
  pages={5131--5148},
  year={2024},
  publisher={Optica Publishing Group}
}

@article{li2024feature,
  title={Feature decoupled knowledge distillation enabled lightweight image transmission through multimode fibers},
  author={Li, Fujie and Yao, Li and Niu, Wenqing and Li, Ziwei and Shi, Jianyang and Zhang, Junwen and Shen, Chao and Chi, Nan},
  journal={Optics Express},
  volume={32},
  number={3},
  pages={4201--4214},
  year={2024},
  publisher={Optica Publishing Group}
}

@article{maqbool2024towards,
  title={Towards optimal multimode fiber imaging by leveraging input polarization and deep learning},
  author={Maqbool, Jawaria and Hasan, Syed Talal and Cheema, M Imran},
  journal={Optical Fiber Technology},
  volume={87},
  pages={103896},
  year={2024},
  publisher={Elsevier}
}

@article{caramazza2019transmission,
  title={Transmission of natural scene images through a multimode fibre},
  author={Caramazza, Piergiorgio and Moran, Ois{\'\i}n and Murray-Smith, Roderick and Faccio, Daniele},
  journal={Nature communications},
  volume={10},
  number={1},
  pages={2029},
  year={2019},
  publisher={Nature Publishing Group UK London}
}

@article{fallahpoor2024deep,
  title={Deep learning techniques in PET/CT imaging: A comprehensive review from sinogram to image space},
  author={Fallahpoor, Maryam and Chakraborty, Subrata and Pradhan, Biswajeet and Faust, Oliver and Barua, Prabal Datta and Chegeni, Hossein and Acharya, Rajendra},
  journal={Computer methods and programs in biomedicine},
  volume={243},
  pages={107880},
  year={2024},
  publisher={Elsevier}
}

@article{yang2023medmnist,
  title={Medmnist v2-a large-scale lightweight benchmark for 2d and 3d biomedical image classification},
  author={Yang, Jiancheng and Shi, Rui and Wei, Donglai and Liu, Zequan and Zhao, Lin and Ke, Bilian and Pfister, Hanspeter and Ni, Bingbing},
  journal={Scientific Data},
  volume={10},
  number={1},
  pages={41},
  year={2023},
  publisher={Nature Publishing Group UK London}
}

@book{goodman2007speckle,
  title={Speckle phenomena in optics: theory and applications},
  author={Goodman, Joseph W},
  year={2007},
  publisher={Roberts and Company Publishers}
}

@article{peng2023histogram,
  title={A histogram-driven generative adversarial network for brain MRI to CT synthesis},
  author={Peng, Yanjun and Sun, Jindong and Ren, Yande and Li, Dapeng and Guo, Yanfei},
  journal={Knowledge-Based Systems},
  volume={277},
  pages={110802},
  year={2023},
  publisher={Elsevier}
}

@article{chung2025histogram,
  title={Histogram-based Res-UNet model for optical sectioning HiLo endo-microscopy},
  author={Chung, Meng-Chen and Chia, Yu-Hsin and Vyas, Sunil and Luo, Yuan},
  journal={Optics Express},
  volume={33},
  number={6},
  pages={12253--12268},
  year={2025},
  publisher={Optica Publishing Group}
}

@article{sadeghi2022histnet,
  title={HistNet: Histogram-based convolutional neural network with Chi-squared deep metric learning for facial expression recognition},
  author={Sadeghi, Hamid and Raie, Abolghasem-A},
  journal={Information Sciences},
  volume={608},
  pages={472--488},
  year={2022},
  publisher={Elsevier}
}

@article{avi2023differentiable,
  title={Differentiable histogram loss functions for intensity-based image-to-image translation},
  author={Avi-Aharon, Mor and Arbelle, Assaf and Raviv, Tammy Riklin},
  journal={IEEE Transactions on Pattern Analysis and Machine Intelligence},
  volume={45},
  number={10},
  pages={11642--11653},
  year={2023},
  publisher={IEEE}
}

@inproceedings{afifi2021histogan,
  title={Histogan: Controlling colors of gan-generated and real images via color histograms},
  author={Afifi, Mahmoud and Brubaker, Marcus A and Brown, Michael S},
  booktitle={Proceedings of the IEEE/CVF conference on computer vision and pattern recognition},
  pages={7941--7950},
  year={2021}
}

@article{viola1997alignment,
  title={Alignment by maximization of mutual information},
  author={Viola, Paul and Wells III, William M},
  journal={International journal of computer vision},
  volume={24},
  number={2},
  pages={137--154},
  year={1997},
  publisher={Springer}
}

@article{chen2020laplacian,
  title={Laplacian pyramid neural network for dense continuous-value regression for complex scenes},
  author={Chen, Xuejin and Chen, Xiaotian and Zhang, Yiteng and Fu, Xueyang and Zha, Zheng-Jun},
  journal={IEEE Transactions on Neural Networks and Learning Systems},
  volume={32},
  number={11},
  pages={5034--5046},
  year={2020},
  publisher={IEEE}
}

@article{xing2021gated,
  title={Gated feature aggregation for height estimation from single aerial images},
  author={Xing, Siyuan and Dong, Qiulei and Hu, Zhanyi},
  journal={IEEE Geoscience and Remote Sensing Letters},
  volume={19},
  pages={1--5},
  year={2021},
  publisher={IEEE}
}

\end{document}